\documentclass{article}
\usepackage{url}
\newcommand{\ber}{\begin{eqnarray}}
\newcommand{\eer}{\end{eqnarray}}

\newcommand{\be}{\begin{equation}}
\newcommand{\ee}{\end{equation}}
\newcommand{\bal}{\begin{align}}
\newcommand{\eal}{\end{align}}
\newcommand{\balnonum}{\begin{align*}}
\newcommand{\ealnonum}{\end{align*}}

\usepackage{times}
\usepackage{authblk}

\usepackage{graphicx}
\usepackage{color}
\usepackage{soul} 
\usepackage{algorithm}

\usepackage{float}
\usepackage{setspace}

\usepackage{amsmath,amssymb}
\usepackage{comment}
\usepackage{epstopdf}
\usepackage[top=1in, bottom=1.25in, left=1.25in, right=1.25in]{geometry}

\global\long\def\spwav{SPARCWave} 
\global\long\def\spscat{SPARCScatter} 
\global\long\def\partition{\mathcal{G}} 
\global\long\def\bcss{\mathcal{D}} 
\global\long\def\signal{\mathbf{x}} 
\global\long\def\weights{\mathbf{w}} 
\global\long\def\wave{\mathbf{v}} 
\global\long\def\wavemat{\mathbf{V}} 
\global\long\def\dwtmatrix{\mathbf{W}} 
\global\long\def\nsignal{n} 
\global\long\def\nclusters{K} 
\global\long\def\ndim{T} 
\global\long\def\nlayers{L} 
\global\long\def\conv{\star} 
\global\long\def\scat{S} 

\global\long\def\tscat{\Delta_t} 
\global\long\def\softthresh{\eta}

\begin{document}

%

%


\title{Clustering Noisy Signals with Structured Sparsity Using Time-Frequency Representation}

\author{ Tom Hope \hspace{1.5cm} Avishai Wagner \hspace{1.5cm} Or Zuk \\
{\it \{tom.hope/avishai.wagner/or.zuk\}@mail.huji.ac.il} }

\affil{ Dept. of Statistics, the Hebrew University of Jerusalem \\
 Mt. Scopus, Jerusalem, 91905, Israel }

\date{}

\maketitle

\begin{abstract}
We propose a simple and efficient time-series clustering framework particularly
suited for low Signal-to-Noise Ratio (SNR),
by simultaneous smoothing and dimensionality reduction aimed at preserving clustering information.
We extend the sparse K-means algorithm by incorporating structured sparsity, and use it to exploit
the multi-scale property of wavelets and group structure in multivariate signals. Finally,
we extract features invariant to translation and scaling with the scattering transform, which corresponds
to a convolutional network with filters given by a wavelet operator, and use the network's structure
in sparse clustering. By promoting sparsity, this transform can yield a low-dimensional
representation of signals that gives improved clustering results on several real datasets.

\end{abstract}

\section{Introduction}

Clustering of high-dimensional signals, sequences or functional
data is a common task that arises in many domains \cite{survey14,liao2005clustering}.
Such data come up in diverse fields, as in speech analysis, genomics, mass spectrometry, MRI or EEG measurements, and many more.

Clustering seeks to partition data into groups with high overall similarity between members (instances) of the same group and dissimilarity
to members of other groups. For time-series signals, this means partitioning the instances into groups of similarly behaving functions over time,
where the measure of similarity is crucial and often application-specific.

In many real-world scenarios, signals are high-dimensional (such as in genomics), noisy (as in low-quality speech recordings), and
exhibit non-stationary behavior: for example peaks and other non-smooth local patterns, or changes in frequency over time.
In addition, signals are often subject to translations, dilations and deformations. These properties generally make clustering difficult.

\begin{figure}[!h]
	\centering
	\small
	\includegraphics[width=4.9in,height = 2.0in]{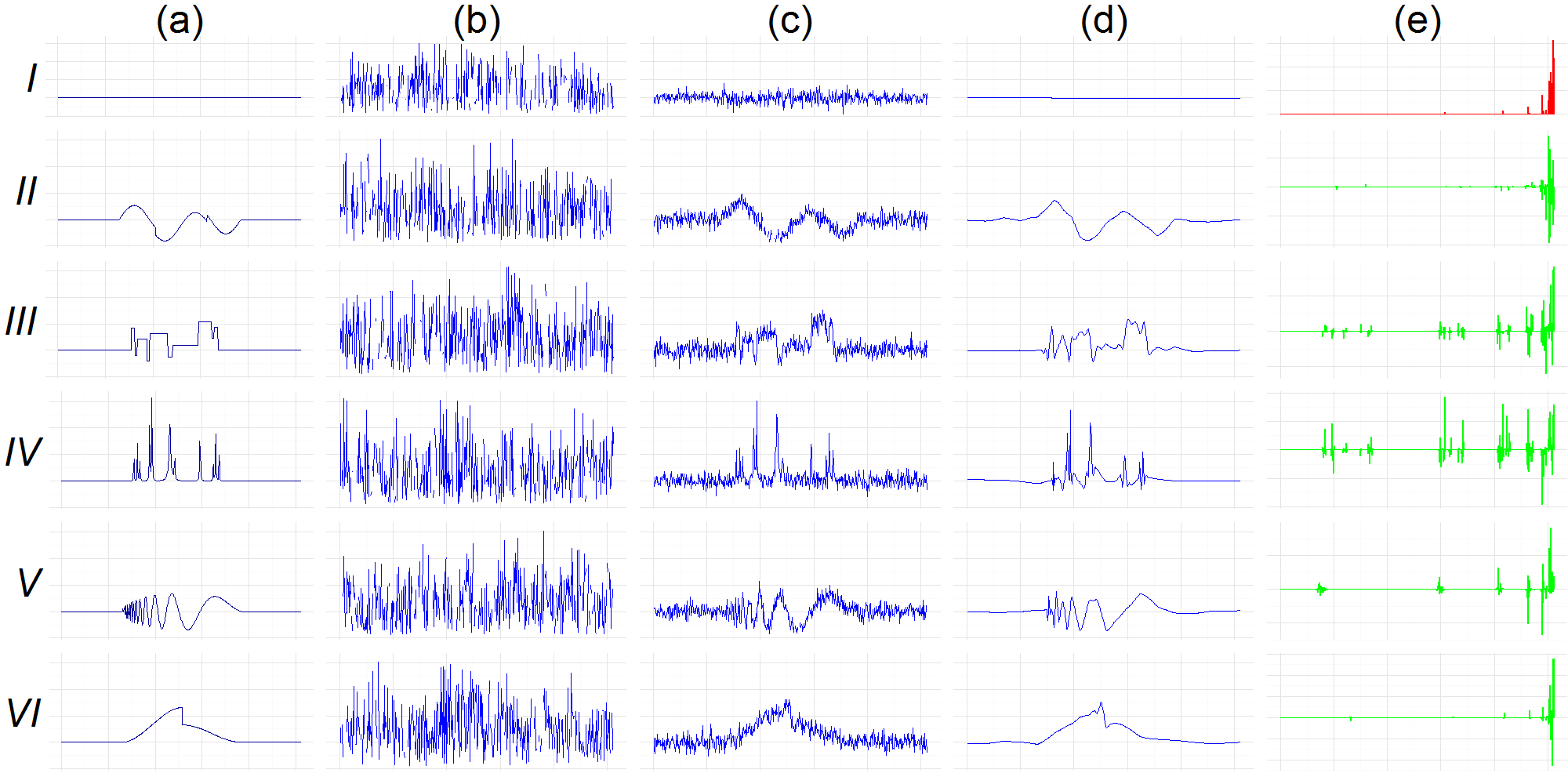} 
	\caption{{\scriptsize \textbf{Simulated cluster data and \spwav's cross-signal smoothing and dimensionality reduction}. Clustering results for six clusters ($I$ - $VI$): Flat curve, Heavisine, Blocks, Bumps, Doppler, Piecewise Polynomial. We generated each individual signal by applying additive Gaussian noise with $\sigma = 2.5$. There are $30$ signals
			for each cluster. We ran \spwav\: as described in the text. \textbf{(a)} True cluster centers. \textbf{(b)} One individual curve from each cluster.
			SNR is too low to allow individual curve smoothing to reliably estimate cluster centers. \textbf{(c)} Cluster centers returned by \spwav\: resemble the original clusters \textbf{(d)} Wavelet smoothing of returned cluster centers improves estimation of original cluster centers.
			\textbf{(e)} Wavelet coefficients for each true cluster center from $II.-VI.$ (except for the trivial flat curve) are ordered from the finest (left) to the coarsest level (right), shown in green. The $w_i$ coefficients fitted in \spwav\: are shown in red. Although a few wavelet coefficients at fine resolutions
			are large for some of the curves, the informative coefficients for clustering are all at the coarsest levels. }}
	\label{fig:sim}
\end{figure}

Clustering is often based on the pairwise distances between signals -
but in the low Signal-to-Noise Ratio (SNR) scenario these distances may be unreliable.
An apparent solution is to first smooth each signal, independently of the others -
but this may be sub-optimal as potentially important clustering information could be lost when noise is too high (see Figure \ref{fig:sim}).
In addition, typical feature representations and distance measures are not invariant to the kind of transformations that occur to real-world signals,
such as translations.

\textbf{Our contributions}. We introduce \spwav \: -- Sparse Clustering with Wavelets -- a framework of methods for sparse clustering of noisy signals,
which \textbf{(i)} uses ``global" cross-signal information for smoothing and dimensionality reduction simultaneously,
by sparse clustering using time-frequency representation, \textbf{(ii)} is geared towards preserving clustering information (rather
than individual signals -- see Figure \ref{fig:sim}), \textbf{(iii)} exploits structure in the signal representation,
such as multi-scale properties in univariate signals and interdependencies in multivariate signals, and \textbf{(iv)} uses the scattering transform
\cite{anden2014deep,bruna2010classification,bruna2013invariant}, that generates features invariant to translations and dilations and robust to small
deformations in both time and frequency. We use the natural structure in the scattering coefficients in our sparse clustering method.

Our methods achieve higher clustering accuracy, compared to methods available in the literature, in simulations on both univariate and multivariate signals,
and on real-world datasets from different domains. We implemented our algorithms and simulations in a python software package ``\spwav" available at
 github \url{https://github.com/avishaiwa/SPARCWave}.

\section{Related Work}
Much work has been done both recently and in the past on using time-frequency representation of signals to cluster them, primarily with wavelets.
Giacofci et al. \cite{giacofci2013wavelet} first perform wavelets-based denoising for
each signal, and then use model-based clustering on the union
set of non-zero coefficients after shrinkage. However, informative features can be thresholded by single-signal denoising due to the lack of global information
highlighting their importance (See Figure \ref{fig:sim}).

Antoniadis et al. \cite{antoniadis2013clustering} use the fact that the overall energy of a signal $\signal \in \mathbb{R}^T$ can be decomposed into a multi-scale view of the energy, and propose extracting features that involve averaging wavelet coefficients in each scale and measuring each scale's relative importance. The averaging of {\it all} wavelets at the same scale in \cite{antoniadis2013clustering} increases SNR and makes the method robust to translations.
However, this averaging loses the time information: two different signals with the same power spectrum will have very similar $\wave_{(j)}$ coefficients,
which may lead to clustering errors.

Non-rigid transformation are typically present for real world signals, making the computed pairwise distances between signals unreliable if these
are not taken into account.
A common approach to dealing with deformations is the Dynamic Time Warping approach (DTW) \cite{berndt1994using}, which uses dynamic programming to
align pairs of signals. However, usage of DTW for our problem is limited by the fact that SNR is typically too low, such that two individual signals
do not contain enough information in order to reliably align them. In our simulations and real-data applications, we did not obtain satisfactory results with DTW.


\subsection{Sparse Clustering}
Several methods for sparse clustering have recently been proposed. For example, in \cite{Wasserman15} the authors consider a
two-component Gaussian Mixture Model (GMM), and provide an efficient sparse estimation method with theoretical guarantees.
While our approach could make use of any sparse clustering method,
we chose sparse K-means \cite{witten2010framework} for its simplicity and ease of implementation.
This simplicity allowed us to more readily incorporate our extensions to structured sparsity for both univariate and multivariate signals.
We use rich feature representations, and hope for linear cluster separability in this feature space.
By using these representations, we are able to achieve good results with the simpler (sparse)
K-means, which corresponds to assuming a spherical GMM.

\section{Methods}
We first provide a brief description of wavelets. Next, we introduce our \spwav \: methods - using sparse K-means in the wavelet domain,
incorporating group structure for univariate and multivariate signals, and finally applying the scattering transform to obtain transformation-invariant features.

\subsection{Wavelets Background}
\label{wave_bckgrnd}
Wavelets are smooth and quickly vanishing oscillating functions, often used to represent data such as curves or signals in time.
More formally, a wavelet family $\Psi_{j,k}$ is a set of functions generated by dilations and translations of a unique mother wavelet $\Psi$:
\begin{equation}
\label{def:wave_funcs}
 \Psi_{j,k}(t) = a^{\frac{j}{2}} \Psi(a^j t - k),
\end{equation}

where $j,k \in \mathbb{Z}$ and $a>0$. The constant $a$ represents resolution in frequency, $j$ represents scale and $k$ translations.
We similarly define a family of functions $\phi_{0,k}$ derived from a scaling function $\phi(t)$ by using the dilation and translation
formulation given in eq. (\ref{def:wave_funcs}).

Any function $f \in L^2(\mathbb{R})$ can then be decomposed in terms of these functions:

\begin{equation}
f(t) = \sum_{k \in \mathbb{Z}} c_{k}\phi_{0,k}(t) + \sum_{j,k \in \mathbb{Z}} \wave_{j,k}\Psi_{j,k}(t)
\label{eq:wavelet_basis},
\end{equation}

where $c_{k} = \langle \phi_{0,k}(t),f(t) \rangle$ are the scaling coefficients, and $\wave_{j,k} = \langle \Psi_{j,k}(t),f(t) \rangle$ are
called the wavelet coefficients of $f(t)$.

This decomposition has a discrete version known as the Discrete Wavelet Transform (DWT). Given $x_1,\ldots, x_T$ forming a signal
$\signal \in \mathbb{R}^T$ sampled at times $t=1,..,T$, where $T = 2^J$, and taking $a=2$, the DWT of $\signal$ is given by
taking $c_0$ and $v_{j,k}$ for $j=0,..,J-1$ and $k=0,..,2^j$ in eq. (\ref{eq:wavelet_basis}),
where $\phi_{0,0}, \Psi_{j,k}$ are evaluated at $t=1,..,T$ to compute the (discrete) inner products
$\langle \phi_{0,0}, \signal \rangle$ and $\langle \Psi_{j,k}, \signal \rangle$.
The DWT can be written conveniently in a matrix form $\wave = \dwtmatrix \mathbf{x}$
where $\mathbf{W}$ is an orthogonal matrix defined by the chosen $\Psi, \phi$, and $\wave$ is the vector of coefficients $c_{0}, \Psi_{j,k}$
properly ordered.

Many real-world signals are approximately sparse in the wavelet domain. This property is typically used for signal denoising
using a three-step procedure \cite{donoho1995noising}: \\
\be
\hat{\signal} \equiv \dwtmatrix^{-1} \softthresh_{\lambda} (\dwtmatrix \mathbf{x})
\label{eq:DWT_denoising}
\ee

where $\softthresh_{\lambda}$ is a nonlinear shrinkage (smoothing) operator, for example entry-wise soft-threshold operator with parameter
$\lambda$: $\softthresh_{\lambda}(\wave) = \text{sgn}(\wave)(|\wave| - \lambda)_{+}$. 

\subsection{Sparse Clustering with Wavelets -- Univariate Signals}

Consider $n$ instances (signal vectors) $\signal^{(i)} \in \mathbb{R}^T$, and let $\wave^{(i)}$ be their DWT transforms.
Let $d_{i_{1},i_{2},j}$ be the squared (Euclidean) distance between $\wave^{(i_1)}$ and $\wave^{(i_2)}$ over coordinate
$j$, $d_{i_{1},i_{2},j} \equiv (\wave_j^{(i_1)}-\wave_j^{(i_2)})^2$.
Let $C_{k}$ be the set of indices corresponding to cluster $k$ with $|C_k|=n_k$. In standard K-means clustering, the goal is to
maximize the \textit{Between Cluster Sum of Squares} (BCSS), which can be represented as a sum
$1^T \bcss$ where $\bcss$ is a vector representing the contribution of the wavelet coefficient $j$ to the BCSS, with

\begin{equation}
	\bcss_j \equiv \frac{1}{n} \sum_{i_{1},i_2=1}^{n} {d_{i_{1},i_{2},j } } - \sum_{k=1}^{K} { \frac{1}{n_{k}} \sum_{i_{1},i_{2} \in C_{k} }{ d_{i_{1},i_{2},j }}}.
\label{eq:BCSS}
\end{equation}

Let $\mathbf{w} \in \mathbb{R}^T$ be a vector of weights, and $s$ a tuning parameter bounding $\|\mathbf{w}\|_1$.
In sparse K-means we maximize a {\it weighted} sum of the contribution from each coordinate, $\mathbf{w}^T \bcss$ \cite{witten2010framework}.
We get the following sparse K-means \cite{witten2010framework} constrained optimization problem, where sparsity is promoted in the wavelet domain:
\begin{equation}
	\begin{aligned}
		& \underset{C_{1},C_{2},...C_{k},\mathbf{w}}{\text{argmax}} \mathbf{w}^T\bcss \\
		& s.t. \quad \mathbf{w} \in \mathcal{B}_2(0, 1) \cap \mathcal{B}_1(0, s) \cap \mathbb{R}^T_{+} \\
	\end{aligned}
\label{eq:sparse_k_means_wavelet_ver2}
\end{equation}

where $\mathcal{B}_p(0, r) \in \mathbb{R}^T$ is the $L_p$ ball centered at zero with radius $r$, and $\mathbb{R}^T_{+}$ is the positive orthant,
$\mathbb{R}^T_{+} \equiv \{ \mathbf{w} \in \mathbb{R}^T, w_j \geq 0, \forall j=1,..,T \}$.

For signals having good sparse approximation in the wavelet domain, we expect that clustering information will also be localized to a few wavelet coefficients,
where high $w_j$ values correspond to informative coefficients (See Figure \ref{fig:sim}).
In \cite{witten2010framework} the authors propose an iterative algorithm for solving Problem (\ref{eq:sparse_k_means_wavelet_ver2}),
where each iteration consists of updating the clusters $C_j$ by running K-means, and updating $\weights$ with a closed-form expression.
The tuning parameter $s$ is set by the permutation-based Gap Statistic method \cite{tibshirani2001estimating}.

\subsection{Group Sparse Clustering}

In many cases, the features upon which we cluster could have a natural structure.
A common example that has seen much interest is group structure, where features come in groups (blocks of pixels in images, neighboring genes, etc.).

Exploiting structured sparsity has been shown to be beneficial in the context of supervised learning (regression, classification),
compressed sensing and signal reconstruction \cite{bach2011convex,huang2011learning}. Here, we extend the idea of using group sparsity to clustering,
and propose the following \textbf{Group-Sparse K-means}.

Let $\partition$ be a partition of $\{1,..,T\}$ and denote by $\wave_{(g)}$ the elements of $w$ corresponding to group $g \in \partition$.
Let $|g|$ the number of elements in $g$. Define the group-norm with respect to the partition $\partition$ \cite{bach2011convex},

\be
|| \wave ||_{\partition} \equiv \sum_{g \in \partition} \sqrt{|g|} || \wave_{(g)} ||_2
\label{def:group_norm}
\ee
where we multiply the vector norms $|| \wave_{(g)} ||_2$ by $\sqrt{|g|}$ as in \cite{chen2012compressive} and \cite{friedman2010note},
to give greater penalties to larger groups (other coefficients may also be used).

Using the group norm, we define the group-sparse K-means problem, where the tuning-parameter $s$ controls group sparsity,
with ${\mathcal{B}_{(\partition)}}(0, s)$ the ball of radius $s$ centered at zero with respect to the group norm $|| \cdot ||_{\partition}$.

\begin{equation}
\begin{aligned}
& \underset{C_{1},C_{2},...C_{K},\weights}{\text{argmax}} \mathbf{w}^T\bcss \\
& s.t. \quad \mathbf{w} \in \mathcal{B}_2(0, 1) \cap {\mathcal{B}_{(\partition)}}(0, s) \cap \mathbb{R}^T_{+}
\end{aligned}
\label{problem_group_sparse}
\end{equation}

Problem (\ref{problem_group_sparse}) generalizes the sparse K-means problem in eq. (\ref{eq:sparse_k_means_wavelet_ver2}),
which is obtained here by setting all groups to have size $1$.

Problem (\ref{problem_group_sparse}) can be solved iteratively as shown in Algorithm \ref{alg:group_sparse_kmeans}.
($\sqrt{\cdot}$ and $\odot$ in step (a.) represent element-wise vector operations: $[\sqrt{\weights}]_j = \sqrt{w_j}$ and $[\weights \odot \wave]_j = w_j v_j$).
Optimization $w.r.t$ $\mathbf{w}$ can be done with standard Second Order Cone Programming (SOCP) solvers by introducing auxiliary variables.
We used the convex optimization toolbox CVX \cite{cvx}.

\begin{algorithm}
	{\bf Input:} $\wave^{(1)},\wave^{(2)},...,\wave^{(n)}$, $K$, $iters$, $\partition$, $s$ \\
	{\bf Output:} Clusters $C_1,C_2,...,C_K$
	\begin{enumerate}
		\item
		\textbf{Initialize}: Set $w_1=w_2=...=w_T=\frac{1}{\sqrt{T}}$
		\item
		\textbf{for} i=$1:iters$
		\begin{enumerate}
			\item
			Apply K-means on $\sqrt{\weights} \odot \wave^{(1)},..., \sqrt{\weights} \odot \wave^{(n)}$
			\item
			Hold $C_1,...,C_K$ fixed, solve (\ref{problem_group_sparse}) w.r.t $\weights$, with tuning parameter $s$ and partition $\partition$
		\end{enumerate}
		\item
		\textbf{Return} $C_1,...,C_K$
		
	\end{enumerate}
	\protect\caption{Group-Sparse K-means}
\label{alg:group_sparse_kmeans}	
\end{algorithm}

We next show two applications of group-sparse clustering in the wavelets domain: (i.) using group sparsity to exploit
structure of the wavelet coefficients, and (ii.) using group sparsity to exploit correlations between different curves in multivariate
signal clustering.

\subsubsection{Exploiting Structure of Wavelet Coefficients using Group Sparsity}

In our time-series clustering context, each scale $g$ of the wavelets coefficients can be thought of as a group,
together forming a multi-scale view of the signal. Since the elements of each group represent ``similar" information
about individual signals, it is natural to expect that they also express similar information with respect to the clustering task.
Therefore, we define the partitions $\partition = \big\{ g_j = \{w_{2^j},...,w_{2^{j+1}-1}\} , j=0,..\log_2(T) \big\}$ with $|g_j| = 2^{j}$,
and solve Problem (\ref{problem_group_sparse}) with the group norm defined by $\partition$.

Our choice of $\partition$ in this work is not the only one possible exploiting the structure of the wavelets coefficients tree -
for example, one may also choose groups corresponding to connected sub-trees (particular nodes and all of their descendent in the tree).
Finally, by allowing partitions $\partition$ with group overlaps, it is possible to get a richer structure for wavelet coefficients, for example by using tree-structured sparsity as in \cite{chen2012compressive}.

\vspace{-0.15cm}

\subsubsection{Clustering Multivariate Signals with Group Sparsity}

\vspace{-0.15cm}
In contrast to the univariate signals discussed so far, in many cases signals are multivariate, and are composed of groups of univariate signals. A simple adaptation of group sparsity enables us to make use of this multivariate structure too.

\begin{figure}
	\centering
	\includegraphics[width=4.4in,height = 2.2in]{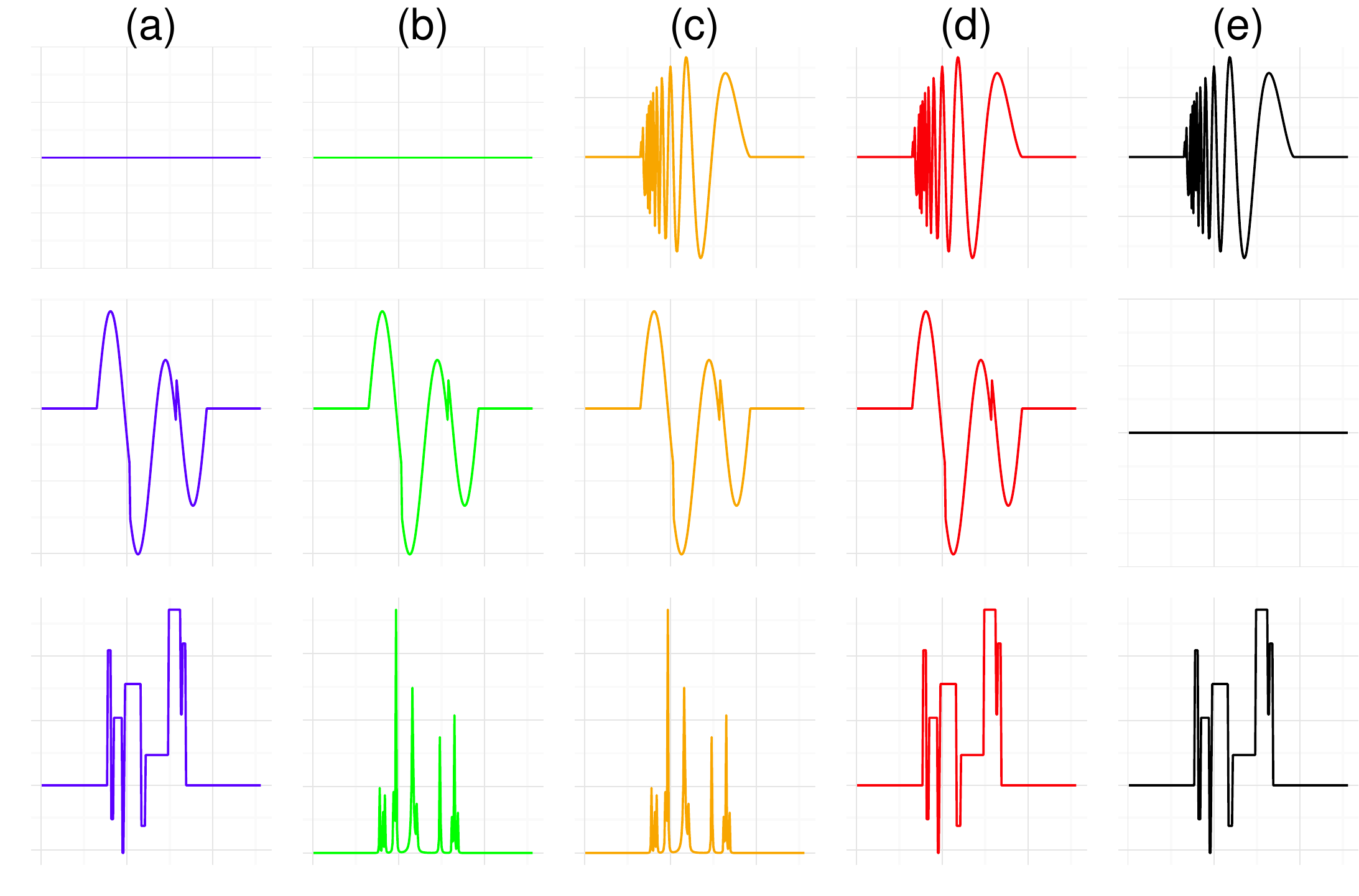}
	\caption{{\scriptsize \textbf{Multivariate cluster data}. Each column (color) is a multivariate cluster center with $G$ variables (here $G=3$).
    Each instance is $\mathbf{X}^{(i)} \in \mathbb{R}^{G\times T}$, with Gaussian additive noise added to each cluster center.}}
	\label{fig:data_multi}
\end{figure}

Let each instance $\mathbf{X}^{(i)}$ be a multivariate signal,	represented as a matrix $\mathbf{X}^{(i)} \in \mathbb{R}^{G\times T}$
where $G$ is the number of variables and $T$ as above (See Figure \ref{fig:data_multi}). We transform each row of $\mathbf{X}^{(i)}$
with DWT to get $\wavemat^{(i)}$. We then concatenate the rows of each $\mathbf{V}^{(i)}$ into one vector of length $G\times T$ and
solve Problem (\ref{problem_group_sparse}) with the group norm $\partition$ defined by
$\partition = \big\{ g_t = \{ {t},{t+T},\ldots,{t+GT}\}, t = 1,..,T \big\}$.
Each group corresponds to a ``vertical time-frequency slice" across the $G$ univariate signals comprising $\mathbf{X}^{(i)}$.

Intuitively, this structure reflects the prior assumption that if at some point $t$ on the time-frequency axis we have an
``active clustering feature", then we expect it to be active across all $G$ univariate signals comprising each $\mathbf{X}^{(i)}$,
but also enables flexibility by allowing $\{w_{t},w_{t+T},\ldots,w_{t+GT}\}$ to be different.

\subsection{The Scattering Transform}
\label{scat}

The wavelet transform is not invariant to translations and scaling. In addition, while the coarse-grain coefficients are stable under small deformations,
the fine resolution coefficients are unstable \cite{mallat2012group}.

To overcome the effects of the above transformations we use the scattering transform \cite{mallat2012group},
a recent extension of wavelets which is stable to such deformations.
The scattering representation of a signal is built with a non-linear, unitary transform computed in a manner resembling
a deep convolutional network where filter coefficients are given by a wavelet operator \cite{anden2014deep}.

More formally, a cascade of three operators - wavelet decomposition, complex modulus, and local averaging - is used as follows.
For a function $f(t)$ the $0$-th scattering layer is obtained by a convolution,
\begin{equation}
\scat^{(0)} \equiv f \conv \phi
\label{eq:scat_layer1}
\end{equation}

where $f \conv g(t) \equiv \int_{s=-\infty}^{\infty} f(s) g(t-s) ds$ and
$\phi$ is an averaging filter.
Applying the filter $\phi$ provides some local translation-invariance but loses high frequency information -
this information is kept by an additional convolution with a wavelet transform described next.

We start by constructing the filters we use in layer $1$. Let $\{\Psi^{(1)}_{j_1}\}_{j_1 \in J_1}$
be a filter-bank containing different dilations of a complex mother wavelet function $\Psi$ for a set of indices $J_1$, with
$\Psi^{(1)}_{j_1}(t) = a_1^{j_1} \Psi(a_1^{j_1} t)$.

To regain translation invariance lost by applying $\Psi^{(1)}_{j_1}$, we perform an additional convolution step with $\phi$ and
obtain the scattering functions of the first layer:
\begin{equation}
\scat_{j_1}^{(1)} \equiv |f \conv \Psi^{(1)}_{j_1}| \conv \phi, \quad \forall j_1 \in J_1
\label{eq:scat_layer2}
\end{equation}

where the absolute value $|\cdot|$ is a contraction operator - this operator reduces pairwise distances between signals,
and prevents explosion of the energy propagated to deeper scattering layers \cite{mallat2012group}.

To recover the high frequencies lost by averaging, we again take convolutions of $|x \conv \Psi^{(1)}_{j_1}|$, with $\Psi^{(2)}_{j_2}$ for some $j_2 \in J_2$.
More generally, define $\Psi^{(l)}_{j_l}(t) = a_l^{j_l} \Psi(a_l^{j_l} t)$ where $a_l$ determines the dilation frequency resolution at layer $l$.
For a scattering layer $l$, let $(j_1,..,j_l)$ define a \textit{scale path} along the scattering network.
This process above continues in the same way, with the scattering functions at the $l$-th layer given by
\begin{equation}
\scat^{(l)}_{j_1,..,j_{l}} \equiv | .. ||f \conv \Psi^{(1)}_{j_1}| \conv \Psi^{(2)}_{j_2}| \conv \Psi^{(3)}_{j_3}| \conv,...,\conv \Psi^{(l)}_{j_l}| \conv \phi,
\: \forall (j_1,..,j_l) \in J_1 \times J_2 \times .. \times J_l.
\label{eq:scat_layerk}
\end{equation}

The corresponding scattering coefficients $v_{j_1,..,j_l}^{(l)}(t)$ are obtained by evaluating $\scat_{j_1,..,j_l}^{(l)}$ at time $t$.

In practice, we let a resolution parameter $\tscat$ determine the points at which we evaluate $\scat^{(l)}_{j_1,..,j_{l}}$,
where for each $\scat^{(L)}_{j_1,..,j_l}$ we only sample at points $t = \frac{k\tscat}{2}$ with $k = 0,\ldots, \lceil\frac{2T}{\tscat}\rceil$.
We stop the process when reaching the final scattering layer $\nlayers$, obtaining scattering coefficients in layers $l=0,1,..,\nlayers$.

We empirically found that in many cases the scattering transform can give a sparse representation of signals,
and that this sparsity helps in the clustering task (see Section \ref{sec:real_data} for real-data results).

\subsubsection{Exploiting Scattering Group Structure}
The scattering coefficients have a natural group structure which we exploit in our clustering.
For each function $\scat^{(l)}_{j_1,..,j_{l}}$, we group all coefficients
resulting from the evaluation of this function. We thus solve Problem (\ref{problem_group_sparse}) with the following definition of groups:
\be
\partition = \big\{ g_{j_1,...,j_l}, \forall l=0,..,\nlayers, \forall (j_1,...,j_l) \in J_1 \times .. \times J_l \big\}
\ee
where
\begin{equation}
g_{j_1,j_2,...,j_l}= \Big\{v_{j_1,\ldots,j_l}^{(l)}(\frac{k\tscat}{2}): k = 0,\ldots, \lceil\frac{2T}{\tscat}\rceil \Big\}
\end{equation}

\section{Simulation Results}
\begin{figure}[H]
	\centering
	\includegraphics[width=4.7in,height = 2.7in]{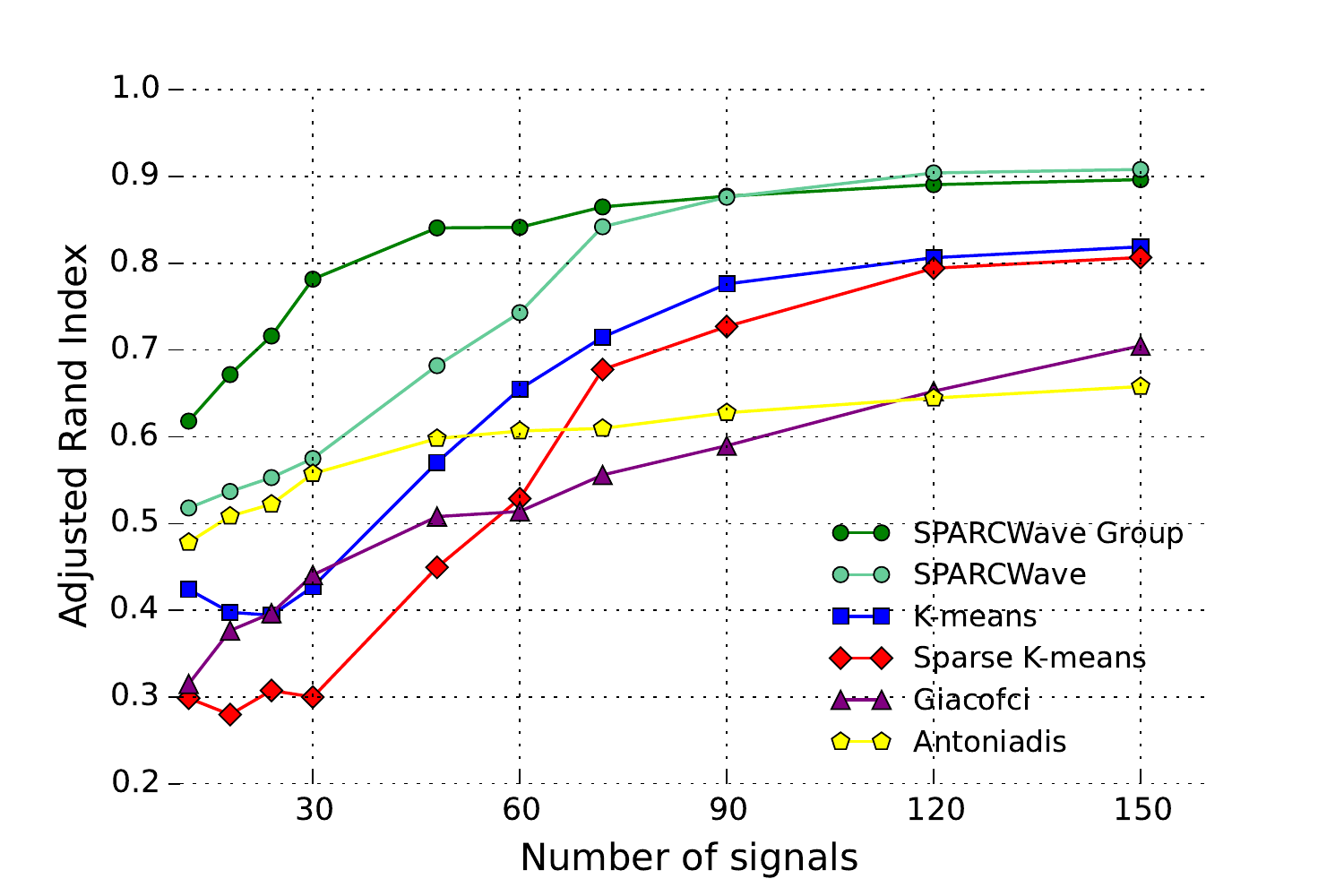}
	\caption{{\scriptsize
\textbf{Univariate simulation results}. Average $ARI$ as a function
of number of signals over $1000$ simulations, for the simulation described in text with $\sigma = 2.75$}
\label{fig:res_single}}
\end{figure}

For univariate signals, we take curves of dimension $256$ from \cite{donoho1995noising} (using the \textit{waveband} R package) and pad them with $128$ zeros on both ends resulting in $K=6$ cluster centers with $T=512$ (see Figure \ref{fig:sim}). Clusters are uniformly sized.
We apply independent additive Gaussian noise $N(0,\sigma^2)$ to generate individual signals. We clustered signals using K-means (picking the best of $100$ random starts), sparse K-means on the raw data as in \cite{witten2010framework} (with the \textit{sparcl} R package, best of $100$ random starts, selecting $s$ using the gap statistic method),
and the methods of \cite{giacofci2013wavelet} (using the \textit{curvclust} R package with burn-in period of $100$) and \cite{antoniadis2013clustering} (using code provided by the authors). For our group-sparse K-means we used CVXPY \cite{cvxpy}, with tuning parameter $s$ selected with the gap statistic method. In all relevant cases, we used the Symmlet-$8$ wavelet.
We used the {\it Adjusted Rand Index} ($ARI$) as a measure of clustering accuracy \cite{wagner2007comparing}. Results are shown in Figure \ref{fig:res_single}.

Group sparsity improved accuracy substantially when the number of curves $\nsignal$ is particularly low (and thus so is SNR).
When $\nsignal$ grows, the more flexible method that does not impose group constraints picks up and the two methods reach about the same accuracy.

\vspace{-0.4cm}
\vspace{-0.4cm}

For multivariate signals, we select curves as shown in Figure \ref{fig:data_multi}. Each univariate signal is of dimension $128$ padded with $64$
zeros on each end, resulting in $K=5$ cluster centers with $G=3,T=256$.
We compared our Group-Sparse K-means method to (i) K-means, (ii) sparse K-means applied to a concatenation of all signals (termed \spwav-Concat)
(iii) a Hidden Markov Model approach in \cite{ghassempour2014clustering} (HMM) (code provided by the authors), and
(iv) a multivariate PCA-similarity measure \cite{yang2004pca} to construct a pairwise distance matrix, which we use in
spectral clustering \cite{ng2002spectral} (PCA) (self-implementation). Group sparsity improved clustering accuracy substantially,
especially for low $\nsignal$.

\vspace{-0.4cm}
\begin{figure}[H]
	\centering
	\includegraphics[width=4.7in,height = 2.7in]{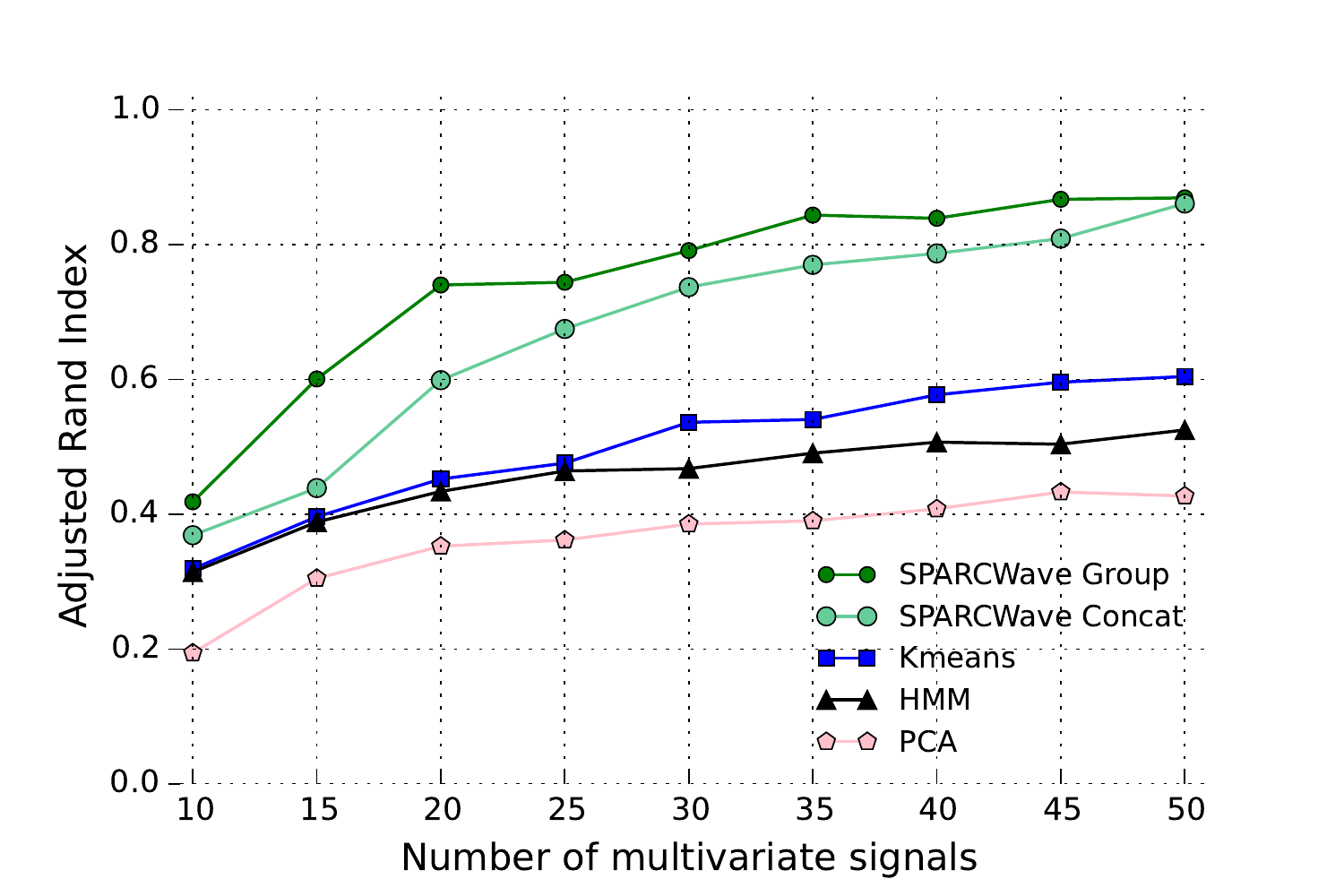}
	\caption{{\scriptsize \textbf{Multivariate simulation results}.
The average $ARI$ is shown as a function of number of signals over $1000$ simulations,
for the simulation described in text with $\sigma = 1.75$. }}
	\label{fig:res_multi}
\end{figure}

\section{Real-Data Applications}
\label{sec:real_data}
\vspace{-0.2cm}

We tested the \spwav\: framework on $3$ real-world datasets. In all cases, using the scattering transform improved clustering accuracy.
We used the implementation in the \textit{scatnet} MATLAB software (available at \url{http://www.di.ens.fr/data/software/}).
We used complex Morlet wavelets for all scattering layers and a moving average for $\phi$.
See the Appendix for specific parameters used in our implementations.
Results for the different methods on all datasets are shown in Table \ref{tab:results_real_data}.

\begin{figure}[H]
	\centering
\includegraphics[width=2.35in,height = 2.7in]{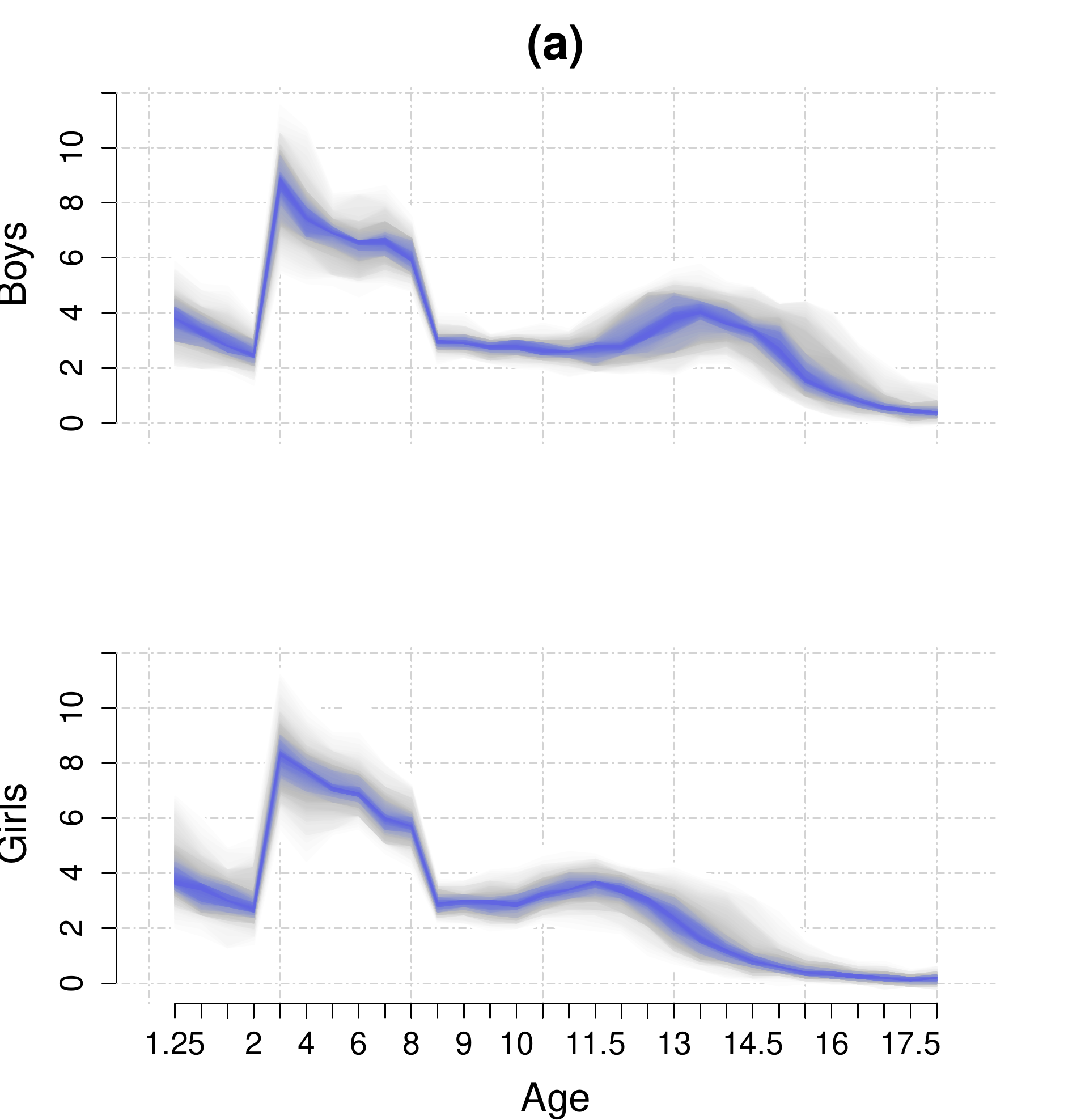}
\includegraphics[width=2.35in,height = 2.7in]{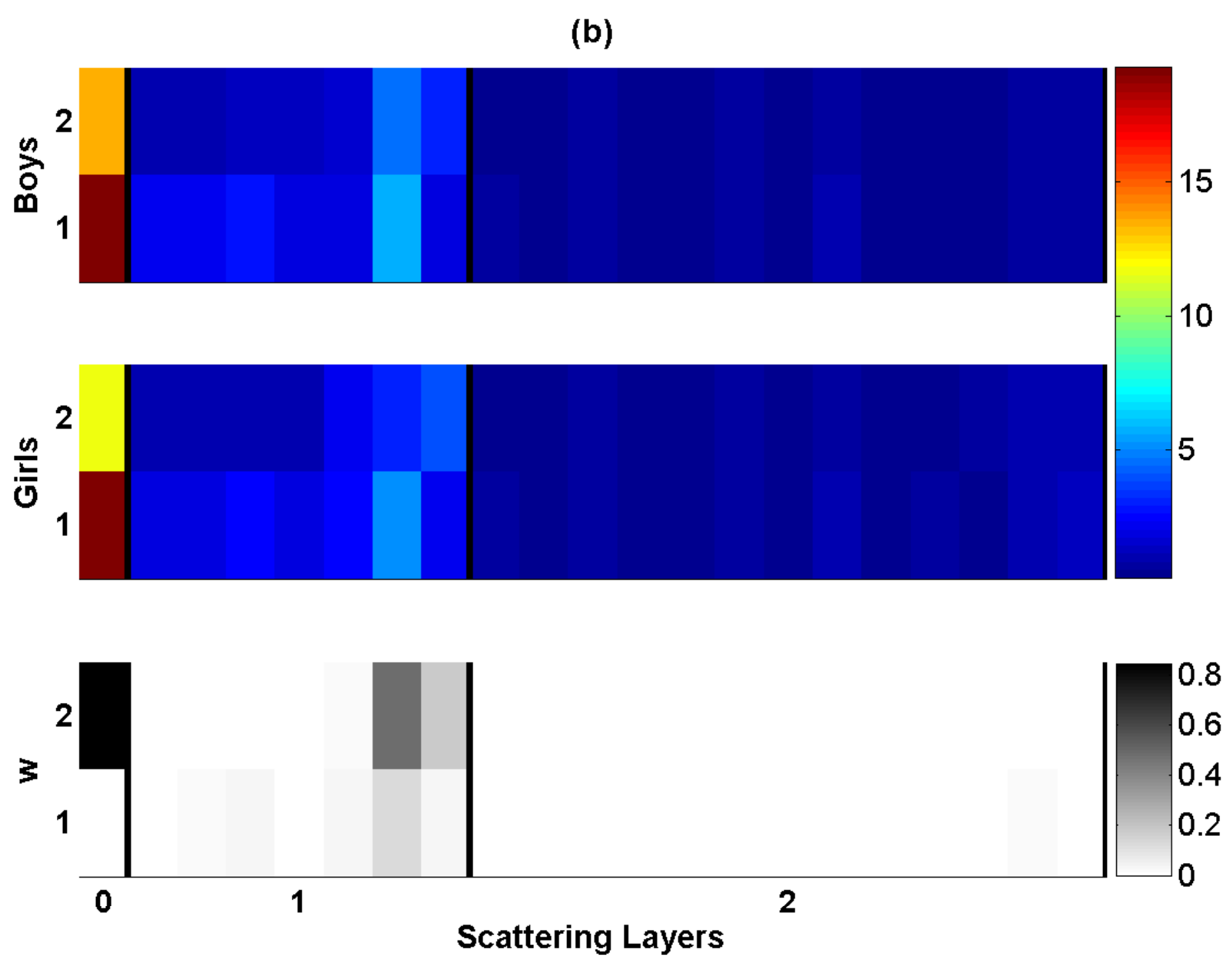}

\caption{{ \scriptsize \textbf{Clustering results for Berkeley Growth dataset}. \textbf{(a)}
Fanplots showing the distribution of first-order differences of the curves in each cluster,
representing the growth rate for boys and girls as function of age. Intra-class variation increases at ages around $12-16$
(indicated by larger faded gray area), motivating the use of the scattering transform which is robust to deformations.
\textbf{(b)} In the first two heat maps, scattering coefficients are shown for the growth rate curves in each cluster,
with colors representing coefficient magnitude. Each row represents a point $t$ at which we evaluate the coefficient
(see Section \ref{scat}). The third heat map represents the \spwav\: weights $\weights$ corresponding to each coefficient.}}
	\label{fig:real}
\end{figure}

\subsection{Berkeley Growth Study}
The Berkeley growth dataset \cite{ramsay2006functional} consists of height measurements for $54$ girls and $38$ boys between the ages of $1$ and $18$
years giving $T=30$ data points for each signal. Of interest to researchers is the ``velocity" of the growth curve, i.e. the rate of change.
We thus use first-order differences of the curves. However, as a result of time shifts among individuals, simply looking at the cross-sectional
mean fails to capture important common growth \cite{tang2008pairwise}, which is also not fully captured by the standard wavelet transform.
We applied the scattering transform together with sparse clustering, which lead to substantial improvements in accuracy.
Applying sparse K-means on the scattering features led to the about the same results as ordinary K-means in this feature space,
with $ARI=0.87$).

The weights $w_j$ selected in the optimization problems (\ref{eq:sparse_k_means_wavelet_ver2}) or (\ref{problem_group_sparse}) may be sub-optimal for clustering.
The features with high-magnitude coefficients can, however, be used in standard K-means, viewing \spwav\: as a feature-selection step \cite{witten2010framework}.
We can select all non-zero features $w_j$ (or above some threshold) and apply K-means using only these features.
This technique improved accuracy from $0.87$ to $0.91$.

Group-sparse clustering with the scattering transform lead to similar results, possibly due to the very short signal length $T$
(see Table \ref{tab:results_real_data} and Figure \ref{fig:real}).

\subsection{Phoneme log-periodograms}
The Phoneme dataset contains $4507$ noisy log-periodograms corresponding to recordings of speakers of $32$ ms duration with $T=256$.
There are $K=5$ phonemes: "sh" ($800$ instances), "dcl" ($757$), "iy" ($1163$), "aa" ($695$), "ao" ($1022$) (See \cite{Tibshirani}
and \url{http://statweb.stanford.edu/~tibs/ElemStatLearn/datasets/phoneme.info.txt}).
Each instance is a noisy estimate of a spectral-density function. De-noising, as suggested for example in \cite{gao1993choice}, is thus attractive here.
By applying the scattering transform on data in the log-frequency domain,
we obtain invariance to translation in log-frequency and consequently to scaling in frequency.
This is similar to the approach taken in \cite{anden2014deep}, where the scattering transform is applied to log-frequencies.
Here too group-sparse clustering did not change accuracy, possibly due to large sample size (see Table \ref{tab:results_real_data}
and Figure \ref{fig:phoneme_real} in the Appendix).

\subsection{Wheat Spectra}

The Wheat dataset consists of near-infrared reflectance spectra of $100$ wheat samples with $T=701$, measured in $2nm$
intervals from $1100$ to $2500nm$, and an associated response variable, the samples' moisture content.
The response variable is clearly divided into two groups: $41$ instances with moisture content lower than $15$,
and $59$ with moisture content above $15$. We use this grouping to create $2$ classes.
Applying the scattering transform in combination with sparse clustering improved accuracy significantly.
However, using group sparsity reduced accuracy, perhaps because out of $924$ features only $2$ are found
to have non-zero weights (see Table \ref{tab:results_real_data} and Figure \ref{fig:wheat_real} in the Appendix).

\begin{table}[!h]
	\caption{Clustering accuracy ($ARI$) for three real datasets. The \spscat\: method shows superior clustering accuracy over
		all datasets.
	}
	\begin{center}
\resizebox{0.8\columnwidth}{!}{%
		\begin{tabular}{|c|ccc|}
            \hline	
			{\bf Method} &{\bf Growth} &{\bf Phoneme} &{\bf Wheat} \\
			\hline
			 \spscat & \textbf{0.91} &\textbf{0.73} & \textbf{0.46} \\
			 SPARCScatter Group & \textbf{0.91} & \textbf{0.73} & {0.3} \\
			 Scattering + K-means & 0.87 & 0.66 & 0.30 \\
			 \spwav & 0.62 & 0.30 & 0.31 \\
			 Antoniadis \cite{antoniadis2013clustering} & 0.62 & 0.34 & 0.35 \\
			 Giacofci \cite{giacofci2013wavelet} & 0.58 & 0.69 & 0.30 \\		
			 K-means & 0.58 & 0.68 & 0.31 \\ \hline
		\end{tabular}
} 
	\end{center}
	
\label{tab:results_real_data}
\end{table}

\vspace{-0.2cm}

\section{Conclusion and Further Work}
We proposed a method for time-series clustering that uses a ``built-in'' shrinkage of wavelet coefficients based on their contribution
to the clustering information. We extended the sparse K-means framework by incorporating group structure and used it to exploit wavelet
multi-resolution properties in univariate signals, and multivariate structure. An interesting future direction is to adapt this
approach to other sparse clustering methods. Another direction is to incorporate richer structures,
such as tree-sparsity in the wavelet and scattering coefficients, long-range dependencies and interdependencies in multivariate signals.
In this work we applied sparse clustering to one-dimensional signals. The wavelets transform
is widely used to represent two-dimensional signals such as images. In addition, a two-dimensional scattering transform achieved
excellent results in supervised image classification tasks due to its invariance to translation, dilation and deformation
\cite{bruna2010classification,bruna2013invariant}. It is thus natural to apply our approach to sparse clustering of images and other multi-dimensional datasets.
Finally, there are few known theoretical guarantees for sparse-clustering methods, and it would be interesting to develop such guarantees to the framework
in \cite{witten2010framework} and it's group extension we have proposed.

\subsubsection*{Acknowledgements} 

We thank Joakim And{\'e}n for suggesting the scattering transform and for many useful comments and discussions.

\newpage
\pagebreak
\clearpage
\begin{spacing}{0.96}
\bibliographystyle{abbrv} 
\bibliography{wavelets_sparse_clustering}
\end{spacing}

\newpage
\clearpage
\onecolumn
\section*{Appendix}

\subsection{Evaluating Performance}
To evaluate the clustering performance we used the {\it Adjusted Rand Index} ($ARI$) as a measure of clustering accuracy \cite{wagner2007comparing},
defined as:
\be
ARI \equiv \frac{\sum_{i < j} \binom{n_{ij}}{2} - \frac{1}{\binom{n}{2}} [\sum_i \binom{a_i}{2}] [\sum_j \binom{b_j}{2}] }
{ \frac{1}{2} [\sum_i \binom{a_i}{2} + \sum_{j} \binom{b_j}{2}] - \frac{1}{\binom{n}{2}} [\sum_i \binom{a_i}{2}] [\sum_j \binom{b_j}{2}] }
\label{def:adjusted_rand_index}
\ee

where $a_i$ denotes the number of signals whose true cluster is $i$, , $b_j$ the number of signals assigned to cluster $j$ and
$n_{ij}$ the number of signals whose true cluster is $i$ and are assigned to cluster $j$, with $n=\sum_i a_i = \sum_j b_j = \sum_{i,j} n_{ij}$.

\subsection{Additional Figures for Simulation}

\begin{figure}[!h]
	\centering
	\includegraphics[width=4.9in,height = 2.0in]{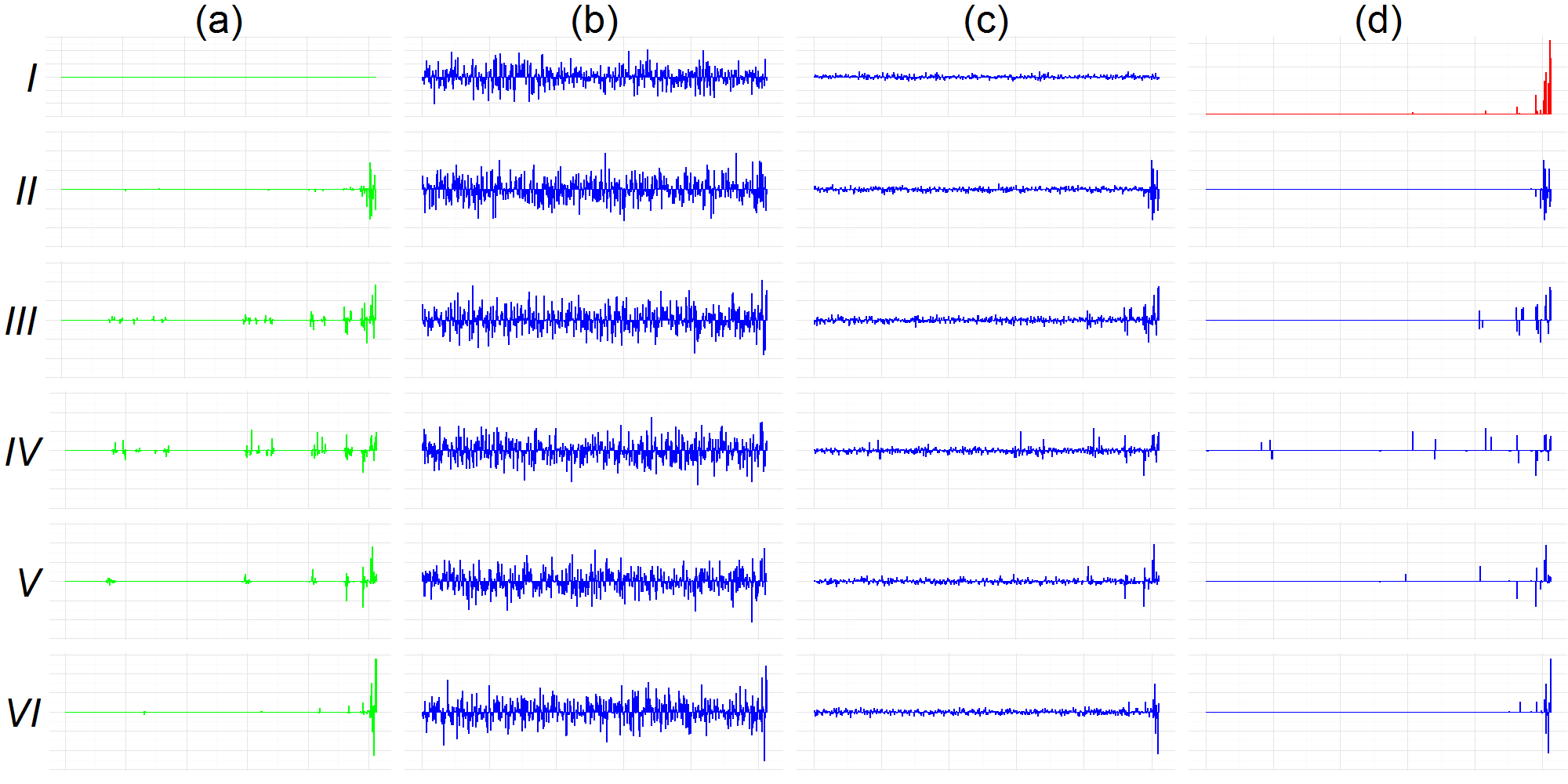}
	\caption{{\scriptsize \textbf{Simulated cluster data coefficients}. Wavelet coefficients for six clusters from the simulation described in main text (see Figure \ref{fig:sim}): $I$. Flat curve, $II$. Heavisine, $III$. Blocks, $IV$. Bumps, $V$. Doppler, $VI$. Piecewise polynomial. \textbf{(a)} Coefficients for true cluster centers.\textbf{(b)} Coefficients for one individual curve from each cluster.
SNR is too low to allow individual curve smoothing to reliably estimate cluster centers. \textbf{(c)} Coefficients of cluster centers
returned by \spwav\: resemble the original clusters coefficients. \textbf{(d)} Coefficients after Wavelets smoothing of returned cluster centers improves estimation of original cluster centers for cluster center from $II.-VI.$ (except for the trivial flat curve), shown in blue. The $w_i$ coefficients fitted in \spwav, shown in red. Although a few wavelet coefficients at fine resolutions
are large for some of the cluster centers, the informative coefficients for clustering are all at the coarsest levels.
	\label{fig:SI_coefficients}}}
\end{figure}

The coefficients of the clustering in simulations are shown in Figure \ref{fig:SI_coefficients}.

\clearpage
\subsection{Additional Details and Figures for Real Data}

\begin{figure}[H]
	\centering
	\includegraphics[width=2.35in,height = 2.95in]{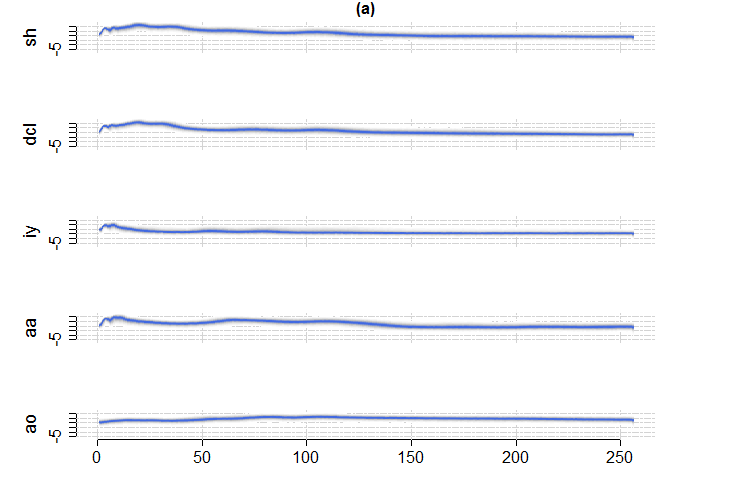}
	\includegraphics[width=2.35in,height = 3in]{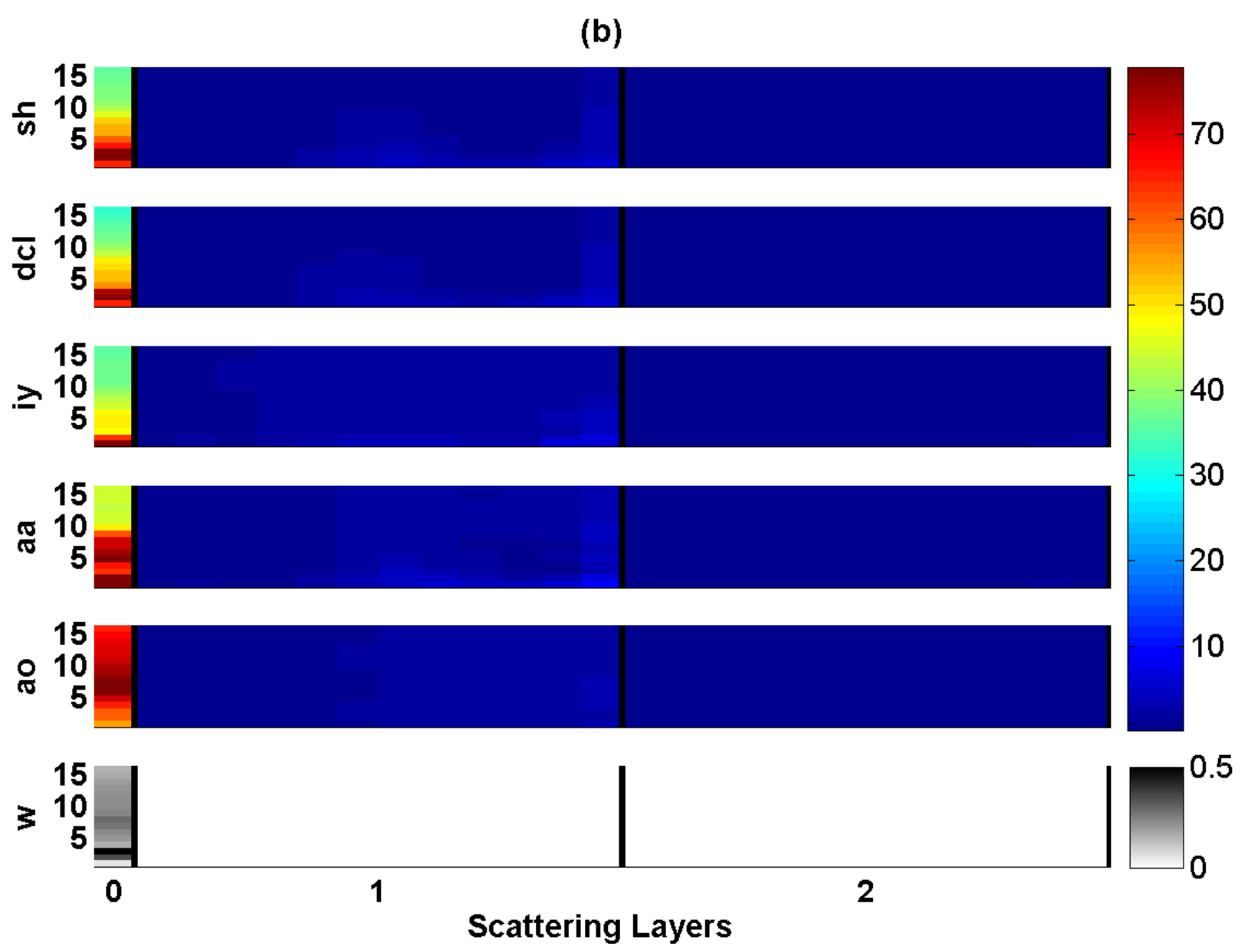}

	\caption{{ \scriptsize \textbf{Clustering results for the Phoneme dataset}. \textbf{(a)}
Fanplots as in Figure \ref{fig:real}. \textbf{(b)} Heat maps for different clusters and for weighs as in Figure \ref{fig:real}.
Most of the clustering information is in the $0$-th layer.}}
	\label{fig:phoneme_real}
\end{figure}

\begin{figure}[H]
	\centering
	\includegraphics[width=2.35in,height = 2.7in]{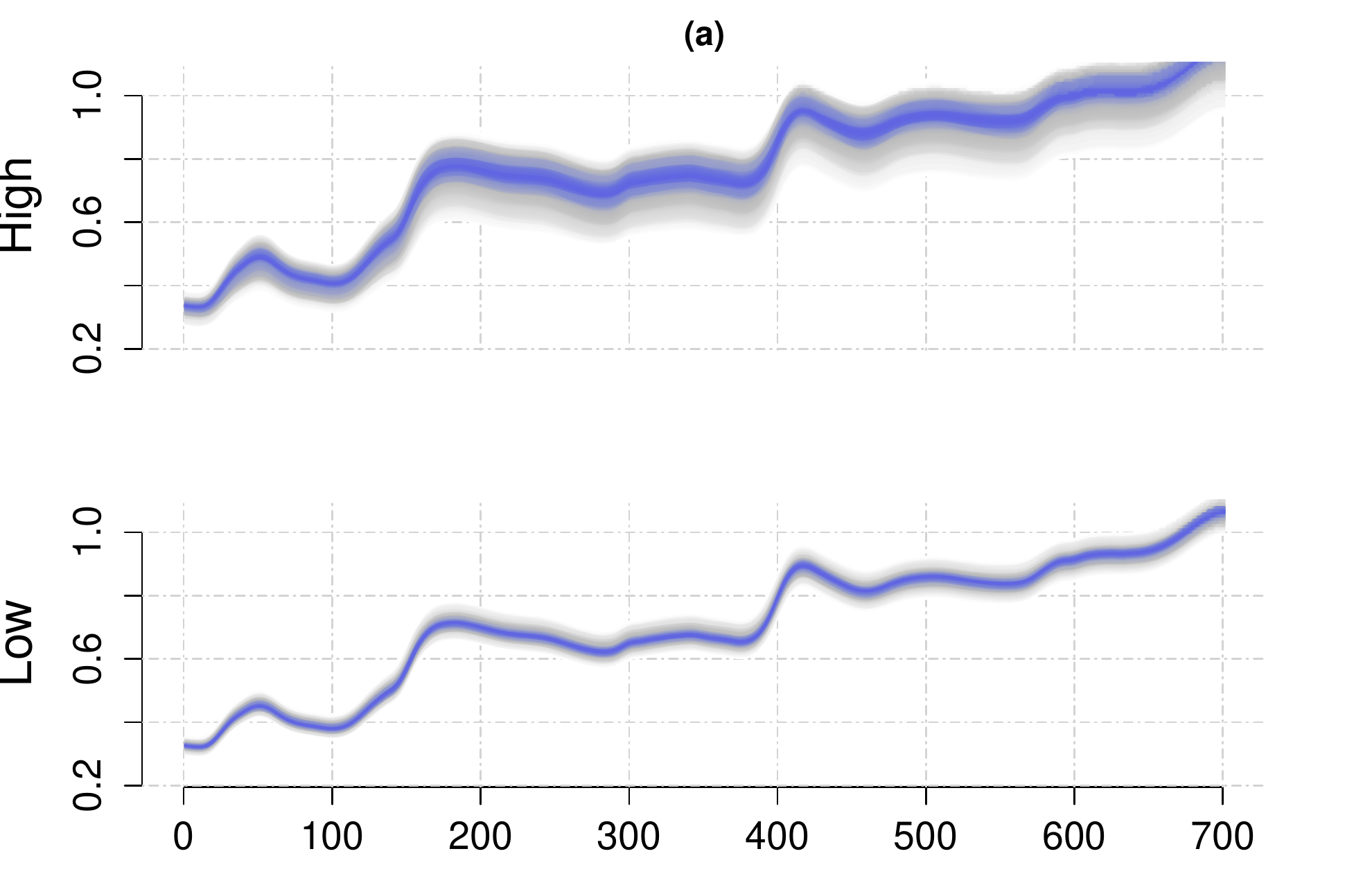}
	\includegraphics[width=2.35in,height = 2.7in]{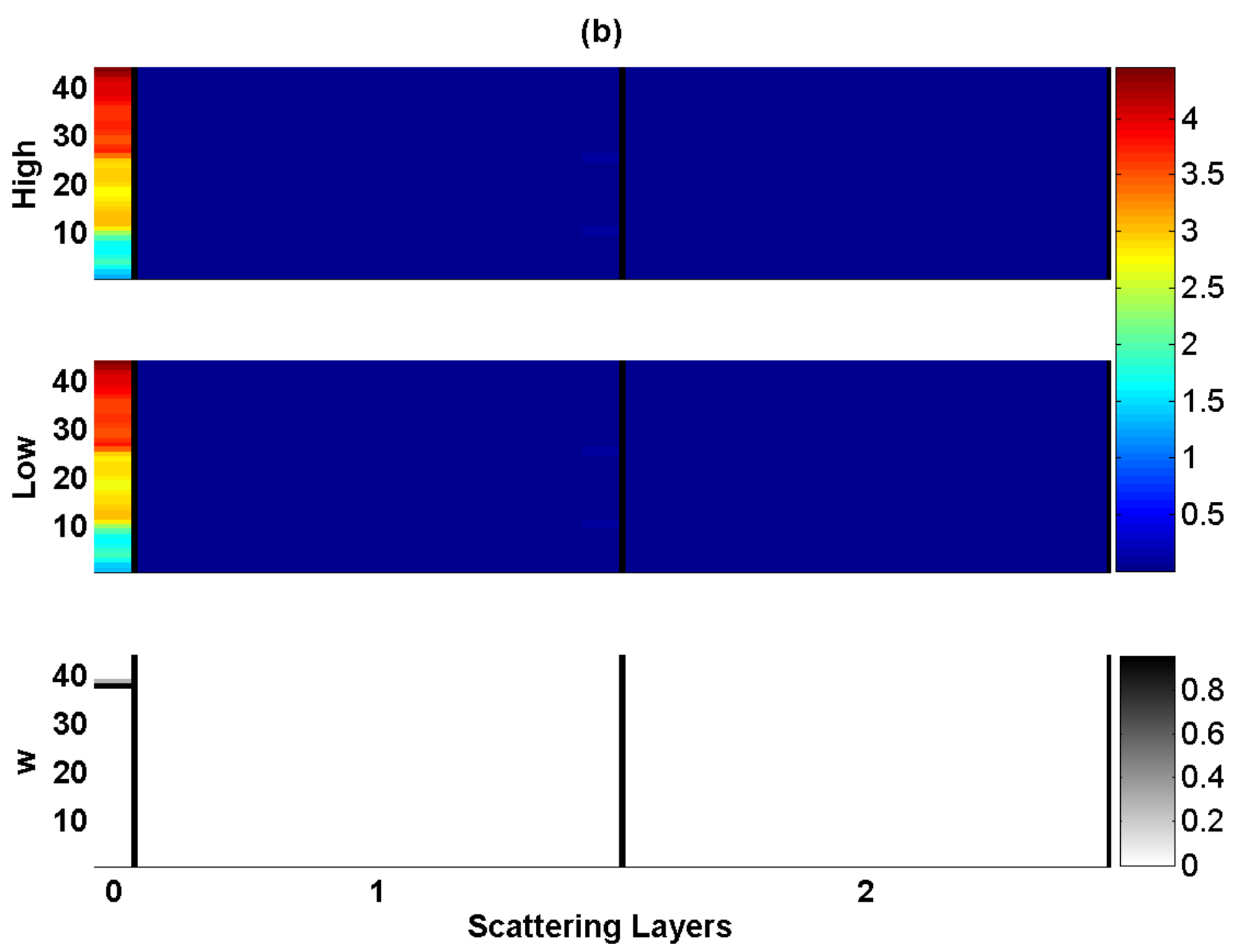}
	
	\caption{{ \scriptsize \textbf{Clustering results for the Wheat dataset}. \textbf{(a)}
Fanplots as in Figure \ref{fig:real}. \textbf{(b)} Heat maps for different clusters and for weighs as in Figure \ref{fig:real}.
Most of the clustering information is in the $0$-th layer.}}
	\label{fig:wheat_real}
\end{figure}

\clearpage
\begin{table}[h]
	\caption{Parameters for running Scattering transform for each of the three real datasets.
We used $\nlayers=2$ in all datasets. The different sizes $\ndim$ of the original vectors lead to different
sizes for the scattering feature vectors $\wave$. In the \textit{scatnet} software implementation,
a few of the scattering coefficients $v_{j_1,..,j_l}^{(l)}(k \tscat)$ are filtered. The choice of the frequency resolutions $a_1$ and $a_2$
were different for different datasets - for audio datasets (such as Phenome), it is known that higher frequency is required to represent the information
in a signal \cite{anden2014deep}, and we used $a_1=2^{1/8}$.
The \spscat\: method selects only a few of the scattering coefficients at the top two layers.}
	\begin{center}
	
\resizebox{0.7\columnwidth}{!}{%
		\begin{tabular}{|c|ccc|}
            \hline	
			{\bf Parameter} &{\bf Growth} &{\bf Phoneme} &{\bf Wheat} \\
			\hline
			 $\nsignal$ & 92 & 4507 & 100 \\
			 $\ndim$ & 30 & 256 & 701 \\
			 $\nclusters$ & 2 & 5 & 2 \\
			 $\nlayers$ & 2 & 2 & 2 \\
			 $|\wave|$ & 42 & 400 & 924 \\
             $||\weights||_0$ & 10 & 16 & 2 \\
			 $\tscat$ & 32 & 32 & 32 \\		
			 $(a_1,a_2)$ & $(2^{1/2}, 2)$ & $(2^{1/8}, 2)$ & $(2^{1/8},2)$ \\ \hline
		\end{tabular}
}
	\end{center}
	
\label{tab:parameters_real_data}
\end{table}

\end{document}